\documentclass[a4paper,twocolumn]{article}
\usepackage{graphicx}   % Replaces epsfig (modern package)
\usepackage{subcaption}
\usepackage{calc}
\usepackage{amssymb}
\usepackage{amstext}
\usepackage{amsmath}
\usepackage{amsthm}
\usepackage{multicol}
\usepackage{times}      % Alternative to pslatex (modern font)
\usepackage{apalike}
\usepackage{algorithm2e}
\usepackage{url}
\usepackage[bottom]{footmisc}
\usepackage{array}
\usepackage[colorlinks=true, linkcolor=blue, citecolor=blue, urlcolor=blue]{hyperref}

\title{Optimizing Retrieval-Augmented Generation of Medical Content for Spaced Repetition Learning}

\author{
    Jeremi I. Kaczmarek\textsuperscript{1,2,3}, 
    Jakub Pokrywka\textsuperscript{2,3}, 
    Krzysztof Biedalak\textsuperscript{4}, \\
    Grzegorz Kurzyp\textsuperscript{3}, 
    Łukasz Grzybowski\textsuperscript{2,5}
}

\date{
    \textsuperscript{1}Poznań University of Medical Sciences, Poland \\
    \textsuperscript{2}Adam Mickiewicz University, Poland \\
    \textsuperscript{3}Wydawnictwo Naukowe PWN, Poland \\
    \textsuperscript{4}SuperMemo World, Poland \\
    \textsuperscript{5}Association for Research and Applications of Artificial Intelligence, Poland \\
    \vspace{1em}
    \today
}

\begin{document}

\maketitle

\abstract{Advances in Large Language Models revolutionized medical education by enabling scalable and efficient learning solutions. This paper presents a pipeline employing Retrieval-Augmented Generation (RAG) system to prepare comments generation for Poland’s State Specialization Examination (PES) based on verified resources. The system integrates these generated comments and source documents with a spaced repetition learning algorithm to enhance knowledge retention while minimizing cognitive overload. By employing a refined retrieval system, query rephraser, and an advanced reranker, our modified RAG solution promotes accuracy more than efficiency. Rigorous evaluation by medical annotators demonstrates improvements in key metrics such as document relevance, credibility, and logical coherence of generated content, proven by a series of experiments presented in the paper. This study highlights the potential of RAG systems to provide scalable, high-quality, and individualized educational resources, addressing non-English speaking users.}

% \onecolumn \maketitle \normalsize \setcounter{footnote}{0} \vfill

\section{\uppercase{Introduction}}

Recent technology development has transformed many sectors, including education, where innovative tools now play a critical role in facilitating the learning process. This paper presents the application of Retrieval-Augmented Generation (RAG) systems in medical education, focusing on optimizing learning for Polish medical specialists by integrating spaced repetition methodologies. The proposed solution merges advanced machine learning algorithms with medical knowledge grounded in Evidence-Based Medicine (EBM), ensuring accuracy and credibility.

Our contribution is presenting a pipeline in which we acquire questions from the State Specialization Examination (PES, \textit{Państwowy Egzamin Specjalizacyjny}) and prepare an online course specifically tailored to aspiring specialists. The course includes RAG-based comments derived from sources within our specialised search engine, which contains high-quality medical content.

Firstly, the questions and sets of possible answers are presented to a user. After a student submits their response, they are shown the correct answer, accompanied by a Large Language Model (LLM)-based explanation and references to medical content. Following a learning session with multiple items, these questions are organized using a spaced repetition algorithm for subsequent review sessions.

Our approach is a scalable, cost-effective system that enhances knowledge retention and minimizes cognitive overload. Our methodology prioritizes generated content verification by humans, leveraging curated materials and collaboration with medical specialists. By employing structured datasets, verified sources, and an optimized pipeline, our system minimizes risks associated with generative models, such as hallucinations \cite{hallucination1}, while ensuring the traceability of information back to authoritative medical sources.

The motivation for this work stems from the pressing need to provide healthcare professionals with tools that facilitate efficient, reliable, and individualized learning experiences. Current medical question banks in Poland often rely on costly expert annotations, making them economically viable only for the most popular exams. By utilizing a RAG-based system, we demonstrate the potential to create high-quality, scalable, and affordable resources that address the specific needs of medical education. This approach represents a significant advancement in making specialized training more accessible and effective for future healthcare providers.

\section{\uppercase{Related Work}}

\subsection{NLP in Medical Education}

New technologies are becoming effective learning tools for medical students. One prominent example is virtual reality (VR) simulations, which enable learners to practice complex procedures and enhance their technical skills in a risk-free environment. Several studies demonstrate that VR-based training improves surgical performance, reduces errors, and shortens procedure times \cite{vr1,vr2,vr3}. Similarly, virtual patient simulators provide interactive scenarios that help medical students develop diagnostic reasoning and clinical decision-making skills \cite{vp1,vp2amboss}.

Another example of a recent educational tool is \textit{ClinicalKey AI} \cite{clinicalkey_ai}, which integrates advanced information retrieval and artificial intelligence to assist clinicians and students in accessing critical medical knowledge. However, developing and implementing comprehensive online learning resources for medical education remains challenging due to inadequate infrastructure, limited faculty expertise, and other barriers \cite{struggles}. Furthermore, ethical considerations are crucial; maintaining transparency, fairness, and responsible technology use is vital for building trust and ensuring patient safety. \cite{ethics}.

In recent years, LLMs specialized in medicine have shown tremendous potential in supporting medical education \cite{medgemini,labrak2024biomistral,OpenMeditron,medllama3v20,OpenBioLLMs,johnsnowlabsmedllama}. These models can serve as virtual assistants, providing immediate feedback on complex questions and generating tailored educational content. Their capabilities are tested against established medical benchmarks, where some have achieved performance levels comparable to or exceeding standard baselines \cite{singhal2022large}.

\subsection{Polish Medical NLP}
LLMs have become a key focus in AI, demonstrating strong performance in processing large datasets and executing natural language processing (NLP) tasks like text generation, translation, and question answering. These capabilities make them promising tools for improving medical practice by enhancing diagnostic accuracy, predicting disease progression, and supporting clinical decisions. By analyzing extensive medical data, LLMs can rapidly acquire expertise in fields like radiology, pathology, and oncology. Fine-tuning with domain-specific literature further enables them to remain current and adapt to different languages and contexts, potentially expanding global access to medical knowledge. However, integrating LLMs into healthcare presents challenges, including the complexity of medical language and diverse clinical contexts that may hinder their ability to fully capture the nuances of practice. Critically, ensuring model fairness and protecting patient data privacy are essential for responsible and equitable healthcare \cite{karabacak2023embracing}.

Several studies have explored the potential of LLMs for Polish-language medical applications. Notably, research has investigated the performance of ChatGPT on the PES across various specialties, including dermatology \cite{lewandowski2023chatgpt}, nephrology \cite{nicikowski2024potential}, and periodontology \cite{camlet2025application}. More extensive research has assessed LLMs performance across all PES specialties \cite{pokrywkapesgpt}. Further work has introduced a new dataset for cross-lingual medical knowledge transfer assessment, comparing various LLMs on the PES, Medical Final Examination (LEK), and Dental Final Examination (LDEK). Additionally, this research has examined LLMs response discrepancies between Polish and English versions of general medical examination questions, using high-quality human translations as a benchmark \cite{grzybowskimedicalexams}. Moreover, a model for automatically parametrizing Polish radiology reports from free text using language models has been proposed, aiming to leverage the advantages of both structured reporting and natural language descriptions \cite{barbaraklaudel}.

\subsection{Spaced Repetition Algorithms}

Spaced repetition is a widely recognized learning technique designed to optimize memory retention through systematic review scheduling. The methodology predicts forgetting curves for individual learners and specific pieces of information, prompting active reviews at the optimal time to minimize the number of repetitions while ensuring a high retention rate. This approach has proven particularly effective for itemized knowledge domains, such as language acquisition, computer science, and medicine.

The concept of spaced repetition in computer-aided learning has been extensively explored since the 1980s when early experiments led to the development of the SM-2 algorithm \cite{wozniak:sm2}. Successive advancements incorporated individualized learning metrics, such as the theory of memory components, which was fully implemented in the SM-17 algorithm \cite{wozniak1995two}. More recently \cite{pokrywka2023modeling}, research demonstrated that applying a negative exponential function as the output forgetting curve, proposed by \cite{wozniak:stability}, significantly enhances the performance of machine learning models such as Long Short-Term Memory (LSTM) networks.

In addition to algorithmic advancements, large-scale machine-learning approaches have been integrated into spaced repetition systems. For example, Half-Life Regression introduced a method to optimize repetition intervals using real-world learning data \cite{duolingo}. Similarly, a Transformer-based model has been employed within a Deep Reinforcement Learning framework to further refine repetition scheduling \cite{transformerspacedrepetition}.

These advancements in spaced repetition methodologies and their application to digital platforms underscore their critical role in high-volume learning tasks, particularly in domains that require robust knowledge retention, such as medical education.

\section{\uppercase{PES Examination}}
\subsection{Examinations for Physicians and Dentists in Poland}

To practice independently and attain specialist certification, physicians and dentists in Poland must, among other requirements, pass specific examinations. These include the LEK (\textit{Lekarski Egzamin Końcowy}, Medical Final Examination), LDEK (\textit{Lekarsko-Dentystyczny Egzamin Końcowy}, Dental Final Examination), LEW (\textit{Lekarski Egzamin Weryfikacyjny}, Medical Verification Examination), LDEW (\textit{Lekarsko-Dentystyczny Egzamin Weryfikacyjny}, Dental Verification Examination), and PES (\textit{Państwowy Egzamin Specjalizacyjny}, State Specialization Examination).\footnote{\url{https://www.cem.edu.pl/egzaminy_l.php}}

LEK and LDEK serve as licensure examinations for domestic graduates, while LEW and LDEW apply to individuals trained outside the European Union. The PES, in contrast, is a certification exam required to attain specialist status in a medical or dental field. Typically, candidates taking the PES have already gained licensure, completed a 12- or 13-month post-graduate internship, and worked as resident doctors in a specialist setting for a period of four to six years. In addition to passing the specialization examination, they must complete mandatory courses, internships, and perform a required set of medical procedures relevant to their discipline to meet all certification criteria. This article focuses on the PES, particularly its written component.

\subsection{PES}

\begin{figure}
    \centering
    \includegraphics[width=1\linewidth]{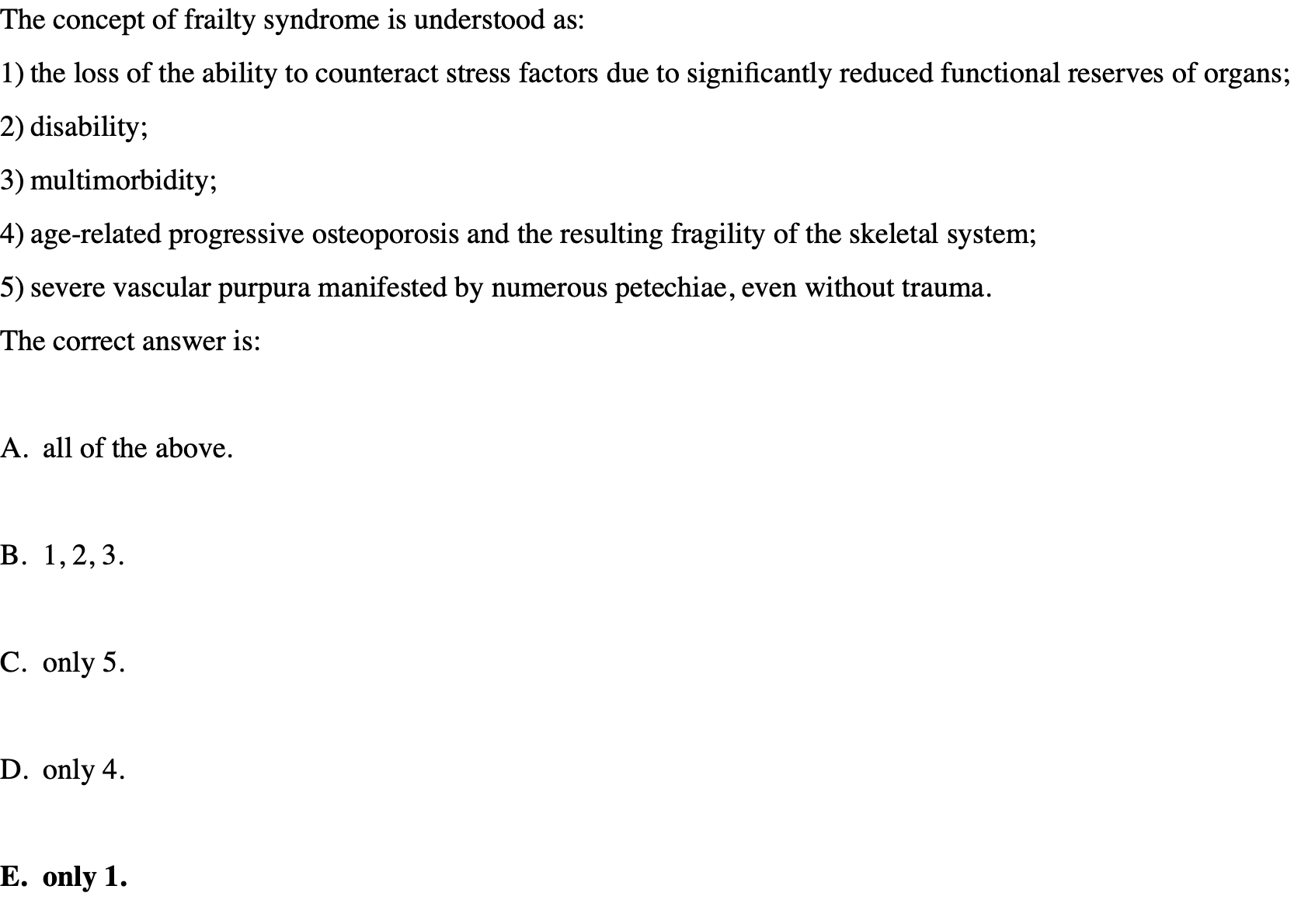}
    \caption{\textbf{PES question.} This example is an English translation of question 10 from the Internal Medicine exam administered during the Fall 2024 session. The translation was prepared by the authors, with the original exam questions written in Polish.}
    \label{fig:enter-label}
\end{figure}

The PES evaluates candidates’ knowledge and competencies to ensure they meet the standards required for specialized practice. It is conducted bi-annually and comprises two main components:

\begin{itemize}
    \item \textbf{Written examination.} This part consists of 120 specialty-specific questions. Each question has five possible answers, with only one correct option. To pass, candidates must achieve a minimum score of 60\%. Since 2022, those scoring 70\% or higher are exempted from the oral examination. The confidentiality of examination questions prior to the test is maintained to ensure the integrity of the process. An example question translated into English is presented in Figure \ref{fig:enter-label}.
    \item \textbf{Oral examination.} Candidates who do not meet the exemption threshold in the written exam participate in the oral component. The structure of this section varies depending on the specialty but typically involves case-based discussions that evaluate clinical reasoning, diagnostic skills, and decision-making abilities.
\end{itemize}

While the oral examination is an integral part of the certification process, this article focuses on the written component due to its relevance to the presented RAG solution. The PES is widely regarded as the most extensive and demanding knowledge verification in the career of a medical professional in Poland.

\section{\uppercase{Learning Platforms Overview}}
In this section, we present existing platforms in which we embedded our PES preparation courses.
% \subsection{Medico}
\subsection{Medico PZWL}

Medical Knowledge Platform Medico PZWL\footnote{\url{https://medico.pzwl.pl/}} is a comprehensive digital resource for Polish doctors, supporting education, clinical practice, and decision-making. Owned by Polish Scientific Publishers PWN\footnote{\url{https://pwn.pl/}}, which has over 80 years of experience in medical education, the platform provides access to an extensive medical knowledge base.

The knowledge base comprises over 120,000 documents, including exclusive content from PWN, as well as materials from other publishers, medical societies, and research institutions. Resources include textbooks, journal articles, clinical guidelines and recommendations, procedural schemes, case studies, surgical records, podcasts, formularies, and legal analyses.

At its core, Medico features an advanced search engine designed to retrieve precise information on specific clinical issues. Search results provide relevant excerpts from various publications, ensuring quick access to the most pertinent insights. This keyword-driven search engine incorporates a reranking mechanism, refined through extensive research and training on over 500,000 expert-annotated medical cases from doctors, paramedics, and medical students\cite{pwnreranking}.

For the development of PES content, RAG queries were initially supplemented with documents retrieved from Medico's production search engine. This retrieval system underwent multiple modifications and enhancements based on feedback from evaluations of the generated content. Since real-time RAG responses were not required, computationally intensive search and reranking improvements were feasible. The final validated comments, along with test materials, were integrated into the SuperMemo spaced repetition application to create an optimized learning experience.

% \begin{figure}[h]
%   \includegraphics[width=0.45\textwidth]{imgs/forgetting_curve_clean.png}
%   % \caption{Forgetting curves applied in learning, source: www.supermemo.com}\label{fig:fcurve}
%   \caption{Forgetting curves applied in learning}\label{fig:fcurve}
% \end{figure}

\subsection{SuperMemo}

SuperMemo\footnote{\url{https://supermemo.com/}} is a world pioneer in applying spaced repetition to computer-aided learning. Its research has been used directly by or inspired the development of this method in other e-learning apps, including Anki, Quizlet or Duolingo. \cite{wozniak:supermemo-history}.

\begin{figure*}[ht!]
    \centering

    % First subfigure
    \begin{subfigure}[t]{0.8\textwidth}
        \centering
        \includegraphics[width=\linewidth]{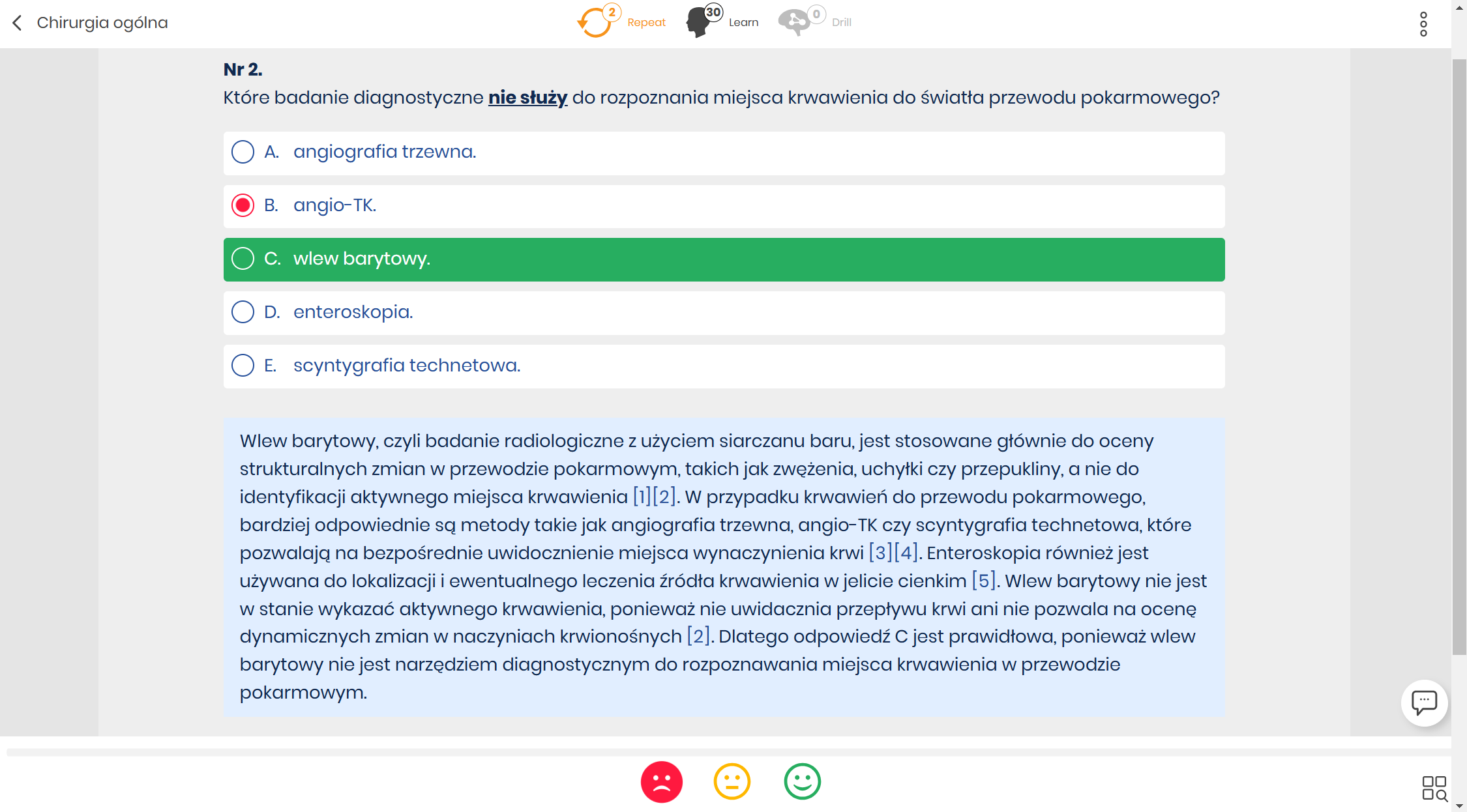}
        \caption{PES multiple choice test with a RAG generated comment in the SuperMemo app. The correct answer (green background) and LLM-generated explanation of a correct answer (blue background) are revealed after the student selects an answer from A, B, C, D, and E choices. The links lead to verified medical documents.}
        \label{fig:pes_comment}
    \end{subfigure}

    \vspace{1em} % Add some space between subfigures

    % Second subfigure
    \begin{subfigure}[t]{0.8\textwidth}
        \centering
        \includegraphics[width=\linewidth]{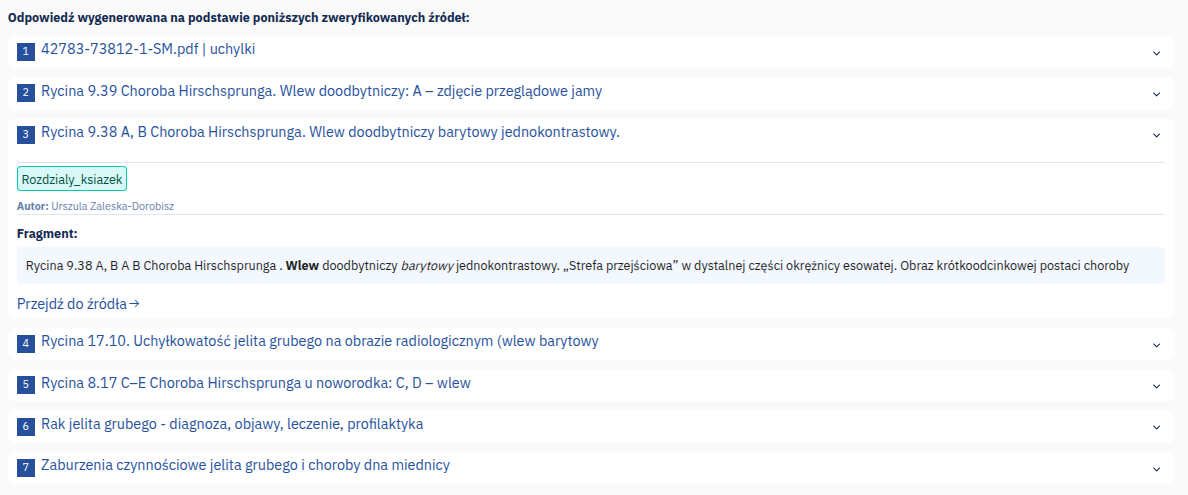}
        \caption{Example sources of LLM-generated information from the medical books, articles, and certified medical websites.}
        \label{fig:sources}
    \label{fig:pescourses}
    \end{subfigure}

    \caption{Overall caption for the figure containing two subfigures.}
    
\end{figure*}

The SuperMemo algorithm consists in predicting forgetting curves individually for each learner and for each information they memorize. Repetitions (active reviews) are planned accordingly in order to minimize the number of them while reaching the desired level of learner's knowledge retention. This is achieved by invoking a repetition of information when its estimated recall probability falls to a required level, typically 90\%. The learner's recall of each piece of information is graded on the scale of "I don't know" :(, "I am not sure or almost right" :|, "I know" :) (see Figure~\ref{fig:pes_comment}). A full history of grades is recorded and used to adapt a general memory model to the individual characteristics of a learner. After each repetition the model is updated and a new forgetting curve of the just reviewed information is estimated. This allows the algorithm to plan the next repetition date. 

Spaced repetition works particularly well for itemized knowledge in areas requiring high-volume learning like languages, computer science, or medicine. The learning content in SuperMemo comprises curated courses as well as memocard (augmented flashcard) collections authored and shared by users. 

\begin{figure}[h!]
  \includegraphics[width=0.45\textwidth]{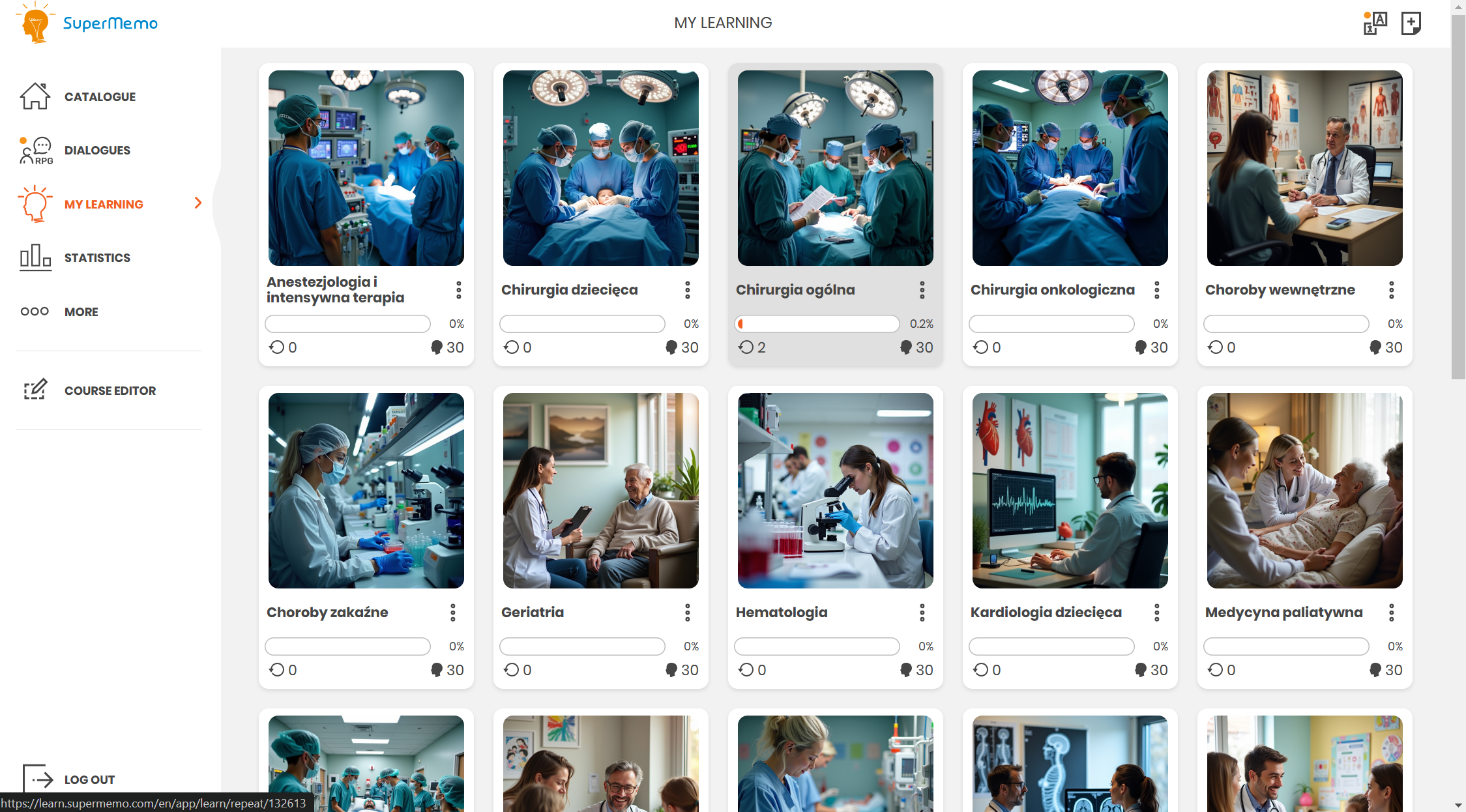}
  \caption{PES courses in the SuperMemo app.}\label{fig:pes_courses}
\end{figure}

\begin{figure*}[h]
  \includegraphics[width=0.8\textwidth]{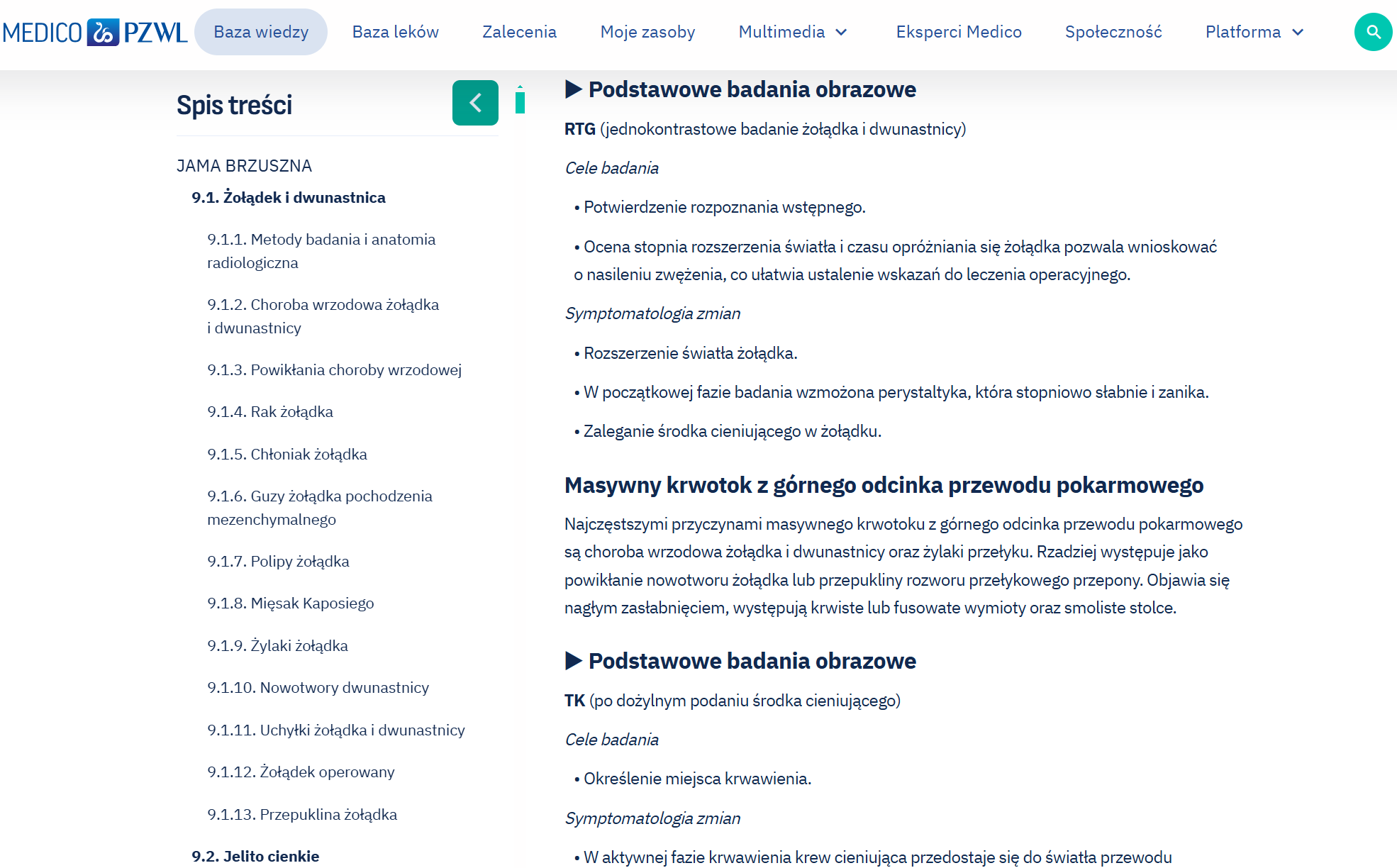}
  \caption{A single source document from Medico called from a link in LLM-generated comment.}\label{fig:medico_source}
\end{figure*}

At the moment of writing this article, the PES courses range in SuperMemo covers actual questions from 4 years of past exams in 22 specializations, all together around 18 thousand items (see Figure~\ref{fig:pes_courses}). Every question is accompanied by a comment generated by a LLM, augmented with a RAG setup supplying relevant source documents from the Medico database (see Figure~\ref{fig:medico_source}). Source documents are quoted and linked in the comments (see Figure~\ref{fig:pes_comment},  and ~\ref{fig:pescourses}).

% \begin{figure*}[h]
%   \includegraphics[width=1.0\textwidth]{imgs/pes-qa-comment-grade.png}
%   \caption{PES multiple choice test with a RAG generated comment. The correct answer (green background) and LLM-generated explanation of a correct answer (blue background) are revealed after the student selects an answer from A, B, C, D, and E choices. The links lead to verified documents.}\label{fig:pes_comment}
% \end{figure*}

% \begin{figure*}[h]
%   \includegraphics[width=1.0\textwidth]{imgs/sources.png}
%   \caption{Example sources of LLM-generated information from medical books, articles, and certified medical websites.}\label{fig:sources}
% \end{figure*}

\section{\uppercase{Exams Aquisition}}
PES exams are publicly released after they are conducted, primarily published on the CEM (\textit{Centrum Egzaminów Medycznych}, Medical Examination Centre) website\footnote{\url{}https://www.cem.edu.pl/} as HTML sites. We employed years 2023 and 2024 part for our learning system. However, the 2021 and 2022 exams were only available as PDFs without a text layer, published by NIL (\textit{Naczelna Izba Lekarska}, Supreme Medical Chamber)\footnote{\url{https://nil.org.pl/}}.

We obtained permission from CEM to use PES materials in our educational platform, ensuring full legal compliance. Below, we describe the process of acquiring and processing these exams.

%\subsection{2021-2022 exams – OCR}
Exams from the years 2021 and 2022 were converted using optical recognition in GPT-4o with JSON mode. A predefined JSON schema structured the extracted data, including test numbers, questions, answer choices, correct answers, and formatting.

%\subsection{2023-2024 exams – WebScrapping}
Tests from the years 2023 and 2024, along with correct answers, were downloaded from the \href{https://cem.edu.pl/index.php}{CEM} website as HTML quizzes using custom Python scraping scripts.

We filtered out items that included visual content like radiological images. Additionally, we removed questions marked as inconsistent with modern medical knowledge. After processing, our dataset comprised 17,843 questions from 149 exams across 46 medical specialties.

\section{\uppercase{Content Generation Pipeline Overview}}

\begin{figure}[h!]
\centering
  \includegraphics[width=0.2\textwidth]{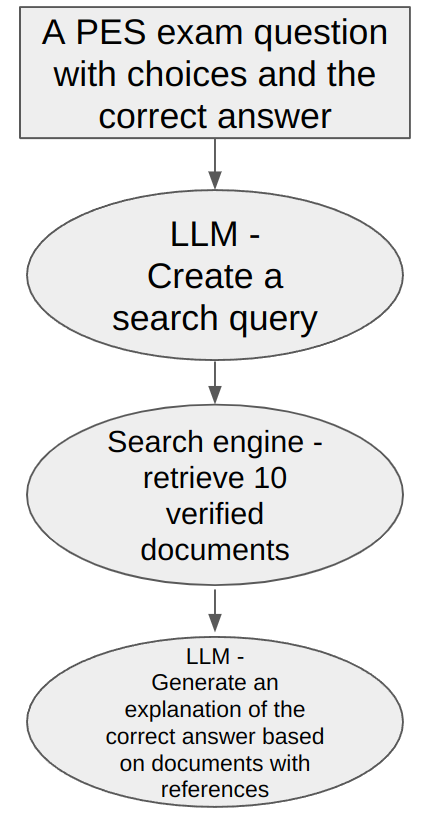}
  \caption{Pipeline overview of selecting relevant documents and generating an explanation of the correct answer. The example of the generated answer is given in Figure \ref{fig:pes_comment} and the example sources are given in Figure \ref{fig:sources}.}
  \label{fig:pipeline}
\end{figure}

Our system, based on the RAG paradigm \cite{rag}, retrieves relevant documents for each PES question and generates concise explanations. Initially, it comprised two core components: a retrieval engine and an answer generation module. During development, we introduced a \textit{Query Rephraser} to enhance search effectiveness — an optional but highly beneficial addition. Each component is detailed in the following subsections, and Figure~\ref{fig:pipeline} illustrates the pipeline. We implemented our system using the Python 3 programming language without any RAG frameworks such as LangChain.

\subsection{Query Rephraser}

Our approach builds on an existing search engine used in real-world medical applications. In production, user queries are typically short (single keywords or a single natural language question) and address a single issue. In contrast, a PES exam questions often consist of multiple sentences, includes several possible answers, and cover multiple topics. Directly inputting the full exam question into the search engine resulted in suboptimal retrieval performance.

To enhance retrieval quality, we introduced a \emph{Query Rephraser}, an LLM-based module that transforms exam questions into optimized search queries. It processes the exam question, answer choices, and the correct answer, generating a concise, targeted query to enhance document retrieval. This query is then processed by our Retrieval System, described in the next subsection.

\subsection{Retrieval System}

Our search engine is a specialized tool for healthcare providers. Its knowledge base consists of authoritative Polish sources authored by medical specialists and researchers, excluding patient-oriented content. Built on Apache SOLR, the retrieval system operates on a keyword-based approach enhanced by an advanced text-analysis pipeline with thousands of domain-specific medical synonyms. Additionally, a cross-encoder-based reranker re-sorts the top 100–200 results. Authors of \cite{pwnreranking} indicated that this approach outperforms a purely bi-encoder-based retrieval pipeline in a medical setting similar to ours.

In production, the search engine must respond within one second, including retrieval and reranking. However, efficiency is not a concern for PES preparatory materials, as courses are generated once, making retrieval quality the priority. To enhance relevance at the cost of computational time, we made several modifications. First, we expanded the reranker’s scope to consider up to 200 candidate documents. Second, we adjusted the reranking context. In the production system, each indexed document corresponds to a book paragraph of approximately 500 words, but the user-facing snippet is limited to about 140 characters. For efficiency and snippet-level optimization, the production reranker relies only on snippet text rather than entire paragraphs. In contrast, for this PES-focused material, we decided to rerank using full paragraphs. We also employed a more powerful reranker, \texttt{dadas/polish-reranker-large-ranknet} \cite{dadas2024assessinggeneralizationcapabilitytext}, which delivers substantially higher quality but is not used in production due to efficiency constraints. These measures significantly improved the relevance of the retrieved documents. Table~\ref{tab:results_table} refers to this refined component as the \emph{Refined Reranker}, whereas the standard production pipeline is denoted as the \emph{Base RAG} system. For implementation, we used the SentenceTransformers library (\url{sbert.net}).

\subsection{Comment Generation}

We chose to employ GPT-4o\cite{GPT4o} via OpenAI API as our large language model, following initial feasibility studies \cite{pokrywkapesgpt,grzybowskimedicalexams} that highlighted its strong performance on the PES exam task, albeit with some remaining inaccuracies. To mitigate this, each question prompt provided the full exam question text, the set of possible answers, and the known correct answer. The model was then instructed to justify the correctness of that answer by leveraging the top 10 retrieved documents.

Identifying an optimal prompt for the LLM required several pilot studies with contributions from both a computer scientist and a medical expert. Ensuring the generated text was of appropriate length and style for medical specialists was crucial. In parallel, we conducted experiments to determine the optimal number of documents to supply. Specifically, we tested a scenario where the LLM relied solely on external documents, disregarding its internal knowledge, and determined the correct answer without prior access to it. We observed that supplying exactly 10 documents yielded optimal results—fewer led to a significant performance drop, while more offered no substantial improvement. This informed our decision to consistently retrieve 10 documents. Final result evaluation, independent of document count, was conducted by specialist annotators, as detailed in the following sections.

\section{\uppercase{Evaluation}}

Improving and validating any system demands reliable methods for assessing its performance. To systematically and objectively evaluate our solution, we created a tailored assessment framework. This approach was consistently applied to successive iterations of the RAG system during its development, enabling comparisons between versions and analyses of how our modifications to the generation pipeline influenced its overall performance.

Polish medical exam preparatory courses often include question banks accompanied by expert-written commentary, which served as our gold standard. Through a thorough analysis of these resources and extensive user consultations, we identified several critical aspects that matter most to physicians. The evaluation model for PES commentary was designed to address the specific needs of the medical field while considering the architecture of the RAG system, including both its inputs and outputs. Below, we present a detailed overview of the framework's structure and the rationale behind its components. A summary of its key assumptions is provided in Table \ref{tab:definition_table}.

\begin{table*}[h!]
    \small
    \centering
    \caption{Parameters utilised by the evaluation framework.}
    \label{tab:definition_table}
    \begin{tabular}{|p{2.75cm}|p{4.75cm}|p{4.75cm}|p{1.75cm}|}
    \hline
        \textbf{Parameter}& \textbf{Definition}& \textbf{Exemplary question}&\textbf{Rating} \\ \hline
         Sensitivity&  The ability to identify actual difficulties in the question.&  Does the model identify all real difficulties in the question?& Scale 1-4\\ \hline 
         Specificity&  Ability to ignore elements of the question that do not constitute difficulties.&  Does the model incorrectly classify elements of the question that are not difficulties as such?& Scale 1-4\\ \hline 
         Completely relevant documents&  Documents that contain all the information necessary to answer the question correctly.&  & Number of documents out of 10.\\ \hline 
         Partially relevant documents&  Documents that contain some, but not all of the information necessary to answer the question correctly.&  & Number of documents out of 10.\\ \hline 
         Relevant documents
(total)&  Documents that contain information necessary to answer the question correctly.&  & Number of documents out of 10.\\ \hline 
         Credibility&  Consistency in citing appropriate sources when they are available.&  For each statement, does the model reference all available relevant sources? Does the model acknowledge using internal knowledge when no suitable sources are available?& Scale 1-4\\ \hline 
         Accuracy&  Authenticity of statements based on paraphrased sources and the model's knowledge.&  Does the model correctly paraphrase statements from the sources it references? Are statements truthful?& Scale 1-4\\ \hline 
         Logic&  Consistency of conclusions drawn in the context of the question with logic.&  Are the conclusions consistent with the cited general statements and logical principles?& Scale 1-4\\ \hline 
         Completeness/Depth&  Comprehensiveness and detail of the explanation.&  Does the commentary address all identified difficulties? Does it sufficiently elaborate on them?& Scale 1-4\\ \hline 
         Conciseness&  Brevity of the commentary.&  Is the commentary overly long? Does it address unnecessary issues?& Scale 1-4\\ \hline 
 Communicativeness/ Readability& Readability and structure of the commentary.& Is the commentary written in a clear and understandable manner?&Scale 1-4\\ \hline 
 Prioritization& Prioritization of higher-value sources in cases of inconsistencies.& Does the model disregard less valuable sources and justify its prioritization of one source over another?&Scale 1-4\\ \hline
    \end{tabular}
\end{table*}

\begin{table*}[ht!]

\centering
\caption{Evaluation results from three experiments on a sample of 200 PES questions across five medical specialties: internal medicine, pediatrics, family medicine, psychiatry, general surgery. A team of five clinical-years medical students assessed the question reports during the development of RAG pipeline. Metric definitions are provided in Table \ref{tab:definition_table}.}
\label{tab:results_table}
\begin{tabular}{|>{\raggedright\arraybackslash}p{5.4cm}|>{\centering\arraybackslash}p{2.85cm}|>{\centering\arraybackslash}p{2.85cm}|>{\centering\arraybackslash}p{2.9cm}|}
\hline
\textbf{Parameter}& \textbf{Base RAG} & \textbf{+Query Rephraser} & \textbf{+Refined Reranker} \\
\hline
Sensitivity (1--4) & \textbf{3.94} ± 0.23 & 3.61 ± 0.56 & 3.76 ± 0.53 \\
Specificity (1--4) & 3.56 ± 0.50 & 3.80 ± 0.44 & \textbf{3.96} ± 0.39 \\
Completely relevant docs (/10) & 2.13 ± 2.39 & 2.75 ± 2.67 & \textbf{3.48} ± 2.73 \\
Partially relevant docs (/10) & 2.46 ± 2.44 & 2.69 ± 2.36 & \textbf{3.35} ± 2.15 \\
Total relevant docs (/10) & 4.59 ± 3.18 & 5.44 ± 2.91 & \textbf{6.83} ± 2.70 \\
Credibility (1--4) & 2.91 ± 0.90 & 2.73 ± 0.84 & \textbf{3.23} ± 0.65 \\
Accuracy (1--4) & 3.46 ± 0.90 & 3.45 ± 0.83 & \textbf{3.56} ± 0.62 \\
Logic (1--4) & 3.67 ± 0.71 & 3.70 ± 0.56 & \textbf{3.86} ± 0.51 \\
Completeness/Depth (1--4) & 3.59 ± 0.67 & 3.73 ± 0.59 & \textbf{3.87} ± 0.37 \\
Conciseness (1--4) & 3.44 ± 0.69 & 3.43 ± 0.62 & \textbf{3.69} ± 0.56 \\
Communicativeness/Readability (1--4) & 3.78 ± 0.31 & 3.80 ± 0.45 & \textbf{3.93} ± 0.30 \\
Prioritization (1--4) & \textbf{3.50} ± 0.84 & 2.60 ± 1.30 & 3.43 ± 0.85 \\
\hline
\end{tabular}
\end{table*}

\subsection{Evaluation Framework}

The approach employs a 1–4 scale to assess various parameters, each describing a certain desired quality. A score of 1 indicates a complete failure to meet expectations, while 4 represents full alignment with criteria, showing no significant shortcomings. Scores of 2 and 3 cover intermediate performance levels, with 2 reflecting predominantly negative aspects and 3 emphasizing mostly positive ones. Neutral scores were excluded to ensure that the annotators adopted a definitive stance. One parameter allowed the annotators to abstain when evaluation was not feasible. Notably, this scale was not used to assess document relevance, which is discussed separately.

\subsubsection{Identification of Key Difficulties}

The first area of the evaluation was LLMs correctness in identifying key difficulties of a question. This was analyzed using two markers: sensitivity and specificity, named after the well-established metrics used in medicine (to evaluate diagnostic test performance) and machine learning (to measure detection effectiveness). Note that unlike their traditional counterparts, these parameters in the described framework are not calculated based on the number of true positives, false positives, true negatives, and false negatives. Instead, they are assessed holistically on a 1–4 scale by human annotators, who evaluate how well the model's responses align with expectations.

\begin{itemize}

\item \textbf{Sensitivity} refers to the model’s ability to identify the true challenges within a question, focusing on essential elements needed to understand and answer it effectively.

\item \textbf{Specificity} evaluates the model’s capacity to disregard irrelevant or superficial details, ensuring it emphasizes meaningful content.

\end{itemize}

There was a risk of neglecting critical aspects or overemphasizing trivial points, undermining commentary quality. Therefore our goal was to ensure that the LLM tailored its output to the users’ level of expertise and underscored the key aspects of the question in a concise comment.

In most iterations, LLM was prompted to identify key difficulties in a given question. Sensitivity and specificity served as metrics to gauge its effectiveness. Alongside other indicators, these metrics helped assess the model’s comprehension of exam questions.

Additionally, identifying key difficulties served as an enabler for the LLM to generate precise search queries and retrieve the most relevant documents, ensuring necessary information was included while irrelevant sources were not. Since the introduction of the \textit{Query Rephraser} module into the generation pipeline, the generated query was evaluated for important and irrelevant elements. Based on this assessment, annotators rated its sensitivity and specificity.

\subsubsection{Document Relevance}

Modern medicine is evidence-based, and consulted physicians emphasized the importance of ensuring that outputs do not contain false or unverified information. To address this, GPT was prompted to generate responses based on source material, with document relevance serving as a key metric for evaluating the quality of the  provided literature. During the evaluation process, we observed that overall appraisal of the comments was most strongly correlated with this parameter. Enhancements that directly increased the number of relevant sources also led to improved scores across all other assessed areas. Annotators noted that when the number of inadequate sources exceeded relevant ones, the comments tended to be vague and included unrelated information, suggesting that GPT prioritized referencing random sources over providing no citations at all. We conclude that document relevance is the most critical indicator of the overall RAG output quality.

Annotators classified documents as:

\begin{itemize}

\item \textbf{Completely Relevant}: Addressed all key difficulties identified by annotators.

\item \textbf{Partially Relevant}: Covered some, but not all, necessary aspects.

\item \textbf{Irrelevant}: Lacked relevance to the question.

\end{itemize}

PES questions often integrate knowledge from multiple areas. For instance, a treatment-focused question may omit a diagnosis, instead presenting symptoms or test results. Answering such questions requires deducing the diagnosis and applying corresponding therapeutic principles. Since relevant information is rarely confined to a single document, a combination of partially relevant sources could prove to be necessary to compile adequate responses.

\subsubsection{Evaluation of Commentary Quality}

The third component of the framework involved a multi-criteria evaluation of commentary quality, assessed using seven parameters rated on a 1–4 scale. These parameters were derived from the expectations of the medical community and underwent extensive consultation processes.

For this evaluation, two types of statements within the commentaries were distinguished: general rules (or factual knowledge), corresponding to textbook theoretical information, and conclusions derived in the context of specific questions. For example, a rule might assert that crushing chest pain is a symptom of myocardial infarction or that changes in serum troponin levels indicate acute myocardial injury, such as during an infarction. Conversely, a conclusion might state that a patient presenting with chest pain should have their troponin levels measured as an element of the myocardial infarction diagnostic process. This distinction was not always straightforward and often depended on contextual factors, including the question's content, possible answers, and source documents.

A detailed description of commentary quality parameters is presented below.

\textbf{Credibility} pertained to factual statements and involved two dimensions. The first dimension was the level of systematic referencing of available sources. The system was expected to cite all documents corroborating the provided information while avoiding references to irrelevant or contradictory content. The second dimension concerned the model's acknowledgement of the use of its internal knowledge. When external sources were incomplete, it was desirable for the system to rely on its internal knowledge while explicitly attributing the information to itself rather than falsely citing external sources or omitting attribution.
Credibility scores were reduced when: 1) a statement lacked references to all corroborating source documents, 2) a statement cited a document that was irrelevant or contradictory. 3) the model failed to attribute internally derived statements to itself.

\textbf{Accuracy} referred to the authenticity of statements, including paraphrases of source documents and the model's internal knowledge. To deem a statement accurate, an annotator needed to identify a corroborating excerpt in at least one document provided to the system or, when this was impossible, validate it through an independent literature review.

\textbf{Logic} was evaluated based on conclusions drawn in the context of the question. This parameter assessed whether the conclusions adhered to logical principles, aligning with the general information cited in the commentary, the question's content, and with each other. Ratings were reduced if conclusions contradicted the cited sources, the correct (non-controversial) answer, or other statements.

\textbf{Completeness/depth} measured the thoroughness and detail of the commentary. It evaluated whether all significant difficulties of the question were addressed and sufficiently detailed explanations were provided to facilitate a full understanding of why one answer was correct and others were not. Ratings were lowered when the commentary failed to address significant issues or addressed them too superficially.

\textbf{Conciseness}, a marker complementary to completeness/depth, assessed the appropriate brevity of the commentary. The system was expected to avoid discussing irrelevant matters or providing excessive detail. Ratings were reduced when commentary included content unrelated to the key difficulties of a question (e.g., irrelevant summaries of source documents) or when the level of detail was excessive from the perspective of the question's requirements.

\textbf{Communicativeness/readability} reflected the linguistic quality and clarity of the commentary, serving as an indicator of grammatical correctness and effective information delivery.

\textbf{Prioritization} evaluated the system's ability to prioritize high-value sources over lower-value ones in cases of conflicting information. Given the rapid evolution of medical knowledge, with new publications rendering older sources obsolete, this parameter aligned with the need for physicians to rely on the most current and reliable evidence. The determination of source value considered factors such as publication date and type of source, with synthesized sources like guidelines, recommendations, and textbooks generally deemed more important than original studies or single-case reports. Since source discrepancies were relatively rare, this parameter was infrequently evaluated, as annotators could abstain from scoring when no contradictions between sources were identified.

\subsection{Evaluation process}

\subsubsection{System development}

During the evaluation of subsequent iterations of the system, a team of five annotators — clinical-years medical students — analyzed and assessed outputs using the established framework. For the evaluation dataset, we selected 40 questions from five exams covering core medical specialties: internal medicine, pediatrics, family medicine, psychiatry, and general surgery. These disciplines were chosen for their broad representation of medical sciences and their relevance to a large proportion of practitioners.

For each question, the assessed output report included:

\begin{itemize}
\item the content of the exam question along with five possible answers,
\item information about the correct answer,
\item a statement by the model containing a list of identified difficulties \textit{or} a single query to the search engine \textit{or} a list of queries to the search engine,
\item 10 source documents,
\item a generated commentary on the question.
\end{itemize}

\subsubsection{Validation}
When output evaluations reached a satisfactory level, we conducted an additional double verification using the same framework to ensure quality control. The validation assessed whether expanding from five core medical specialties to 22, including narrower fields, affected commentary quality. This concern arose because PES questions in narrow specialties are often more detailed and nuanced, with less available relevant content. The validation also enabled inter-annotator agreement comparison and provided insights into the objectivity and reliability of the evaluation methodology.

Unlike during development, validation included 10 questions from each of 22 specialties (9 from emergency medicine, as one question contained an image unsuitable for automatic comment generation). Additionally, a second team of six final-year medical students from a different university joined the annotation process. During validation, 219 comments and 2,190 source documents were evaluated by one member from each team, meaning that every question report was independently reviewed by two unrelated annotators.

Discrepancies were resolved by a third annotator, who reviewed the question report for the first time solely to settle disagreements. A discrepancy was defined as:

\begin{itemize}
    \item One annotator marking a document as irrelevant, while the other considered it partially or fully relevant.
    \item One annotator assigning a score of 1 or 2, while the other assigned 3 or 4.
    \item One annotator deeming prioritization assessable, while the other did not.
\end{itemize}

Annotations following the same general tendency were not considered discrepancies. If one annotator assigned a score of 1 and the other 2, or one rated a document fully relevant while the other deemed it partially relevant, these cases were classified as partial inter-annotator agreement (PIAA) and did not undergo third-party resolution. The resolving annotator assessed only the conflicting ratings, leaving total inter-annotator agreement (TIAA) and PIAA unchanged, and was aware of prior disagreements.

\begin{table}
    \centering
    \begin{tabular}{|>{\raggedright\arraybackslash}p{0.37\linewidth}|>{\centering\arraybackslash}p{0.215\linewidth}|>{\centering\arraybackslash}p{0.095\linewidth}|>{\centering\arraybackslash}p{0.095\linewidth}|} \hline  
         \textbf{Parameter}&  \textbf{Score}&  \textbf{TIAA}&  \textbf{PIAA}\\ \hline  
         Relevant docs (/10)&  6.11 ± 2.91&  57\%&  18\%\\ \hline  
         Credibility (1–4)&  2.92 ± 0.72&  38\%&  28\%\\ \hline  
         Accuracy (1–4)&  3.57 ± 0.66&  57\%&  32\%\\ \hline  
         Logic (1–4)&  3.68 ± 0.46&  58\%&  28\%\\ \hline  
         Completenes/Depth (1–4)&  3.64 ± 0.49&  55\%&  32\%\\ \hline  
         Conciseness (1–4)&  3.63 ± 0.48&  58\%&  31\%\\ \hline  
         Communicativeness/ Readability (1–4)&  3.71 ± 0.36&  58\%&  34\%\\ \hline  
         Prioritization (1–4)&  3.78 ± 0.63&  90\%&  0\%\\ \hline 
    \end{tabular}
    \caption{\textbf{Validation results and inter-annotator agreement statistics.} The table presents the mean scores (± standard deviation) for each evaluated parameter, along with the percentage of total inter-annotator agreement (TIAA) and partial inter-annotator agreement (PIAA). TIAA represents instances where two independent annotators assigned identical ratings, while PIAA reflects cases where ratings followed the same general tendency but were not identical.}
    \label{tab:validation_table}
\end{table}

Final validation results, based on TIAA, PIAA, and discrepancies resolved by a third annotator, along with inter-annotator agreement statistics, are presented in Table \ref{tab:validation_table}. Sensitivity and specificity were not evaluated. The total number of relevant documents and credibility scores were noticeably lower than in the final pipeline evaluation, possibly due to limited relevant content in narrow medical fields. Other parameters retained their values from the end of development. We suggest that the drop in credibility with preserved accuracy could mean that the model compensated for the lack of relevant sources with its internal knowledge. This aligns with annotators' observations, as they did not detect increased factual errors but noted a decline in proper source attribution, reflected in the credibility metric.

Most parameters had a TIAA above 50\%, with overall agreement (TIAA + PIAA) typically reaching 80–90\%. Credibility showed lower inter-annotator agreement, with a TIAA of 38\% and a TIAA + PIAA of 66\%, indicating a need for more precise evaluation guidelines. TIAA for prioritization was 90\%, with the remaining 10\% of discrepancies solely related to assessability. When both annotators deemed it evaluable, their ratings were identical.

Overall, annotators agreed in most cases, with complete disagreements being clear but infrequent. These results highlight the evaluation framework’s potential as a universal tool for RAG development in the medical domain. However, further refinements and improved standardization are desirable to enhance objectivity and repeatability.

\section{\uppercase{Conclusions}}

In this paper, we introduced a pipeline for generating LLM-based content tailored for medical specialists. The content is enriched with verified medical documents and made accessible through our platform. Exam questions are seamlessly integrated into a learning system that employs a spaced repetition algorithm to optimize knowledge retention.

Our approach prioritizes content relevance over efficiency, distinguishing it from typical RAG-based systems. Key enhancements include a \textit{Query Rephraser}, an advanced retrieval system, and a refined reranker. These improvements significantly increased retrieval performance, notably raising the number of total relevant documents from 4.59 to 6.83 out of 10.

To ensure quality and reliability, the output underwent rigorous manual verification by medical specialists. The system is now in its final development stages and will soon be deployed in production.

\section*{\uppercase{Acknowledgements}}
We acknowledge CEM's courtesy in permitting the use of past PES questions and extend our gratitude to all medical annotators for their contributions to development and validation.
This article has been written with the help of GPT-4o \cite{GPT4o} and Grammarly AI Assitant \cite{grammarly}.

\bibliographystyle{apalike}
{\small
\bibliography{example}}

\end{document}